\documentclass[10pt,twocolumn,letterpaper]{article}

\usepackage{iccv}
\usepackage{times}
\usepackage{epsfig}
\usepackage{graphicx}
\usepackage{amsmath}
\usepackage{amssymb}

\usepackage{booktabs}
\usepackage{color}
\usepackage{subfig}

\usepackage{multirow}
\usepackage[export]{adjustbox}
\usepackage{threeparttable}
\usepackage{subfloat}
\usepackage{tabularx}
\usepackage{adjustbox}

\usepackage{multirow}
\definecolor{linkcolor}{RGB}{255,0,0}
\definecolor{urlcolor}{RGB}{255,105,180}
\definecolor{citecolor}{RGB}{66,168,235}

\newcommand{\Q}{\boldsymbol{Q}}

\definecolor{linkcolor}{RGB}{255,0,0}
\definecolor{urlcolor}{RGB}{255,105,180}
\definecolor{citecolor}{RGB}{66,168,235}
\usepackage[pagebackref=true,breaklinks=true,colorlinks=true,bookmarks=false,linkcolor=linkcolor,urlcolor=urlcolor,citecolor=citecolor]{hyperref}

\usepackage[capitalize]{cleveref}
\crefname{section}{Sec.}{Secs.}
\Crefname{section}{Section}{Sections}
\Crefname{table}{Table}{Tables}
\crefname{table}{Tab.}{Tabs.}

\iccvfinalcopy 

\def\iccvPaperID{5110} 

\ificcvfinal\fi

\begin{document}

\title{Betrayed by Captions: Joint Caption Grounding and Generation for Open Vocabulary Instance Segmentation}

\author{
Jianzong Wu$^{1}$\thanks{The first two authors contributed equally to this work. \textsuperscript{$\dagger$} Corresponding Author and Leader. Code and model are available at \url{https://github.com/jianzongwu/betrayed-by-captions}. } \quad
Xiangtai Li$^{2*}$ \textsuperscript{$\dagger$} \quad
Henghui Ding$^{2}$ \quad
Xia Li$^{3}$ \quad \\
Guangliang Cheng$^{4}$ \quad 
Yunhai Tong$^{1}$ \quad
Chen Change Loy$^{2}$
\\[0.1cm]
\small $ ^1$ Key Laboratory of Machine Perception, MOE, School of Artificial Intelligence, Peking University \\
\small $ ^2$ S-Lab, Nanyang Technological University \quad $ ^3$ ETH Zurich \quad
\small $ ^4$ SenseTime Research \\
{\tt\small jzwu@stu.pku.edu.cn~~\{xiangtai.li, henghui.ding, ccloy\}@ntu.edu.sg}
}

\maketitle

\begin{abstract} 
In this work, we focus on open vocabulary instance segmentation to expand a segmentation model to classify and segment instance-level novel categories. Previous approaches have relied on massive caption datasets and complex pipelines to establish one-to-one mappings between image regions and words in captions. However, such methods build noisy supervision by matching non-visible words to image regions, such as adjectives and verbs. 
Meanwhile, context words are also important for inferring the existence of novel objects as they show high inter-correlations with novel categories.
To overcome these limitations, we devise a joint \textbf{Caption Grounding and Generation (CGG)} framework, which incorporates a novel grounding loss that only focuses on matching object nouns to improve learning efficiency.
We also introduce a caption generation head that enables additional supervision and contextual modeling as a complementation to the grounding loss.
Our analysis and results demonstrate that grounding and generation components complement each other, significantly enhancing the segmentation performance for novel classes. 
Experiments on the COCO dataset with two settings: Open Vocabulary Instance Segmentation (OVIS) and Open Set Panoptic Segmentation (OSPS) demonstrate the superiority of the CGG.
Specifically, CGG achieves a substantial improvement of \textbf{6.8\% mAP} for novel classes without extra data on the OVIS task and \textbf{15\% PQ} improvements for novel classes on the OSPS benchmark.
\end{abstract}
\section{Introduction}

\begin{figure}[t!]
	\centering
	\includegraphics[width=1.0\linewidth]{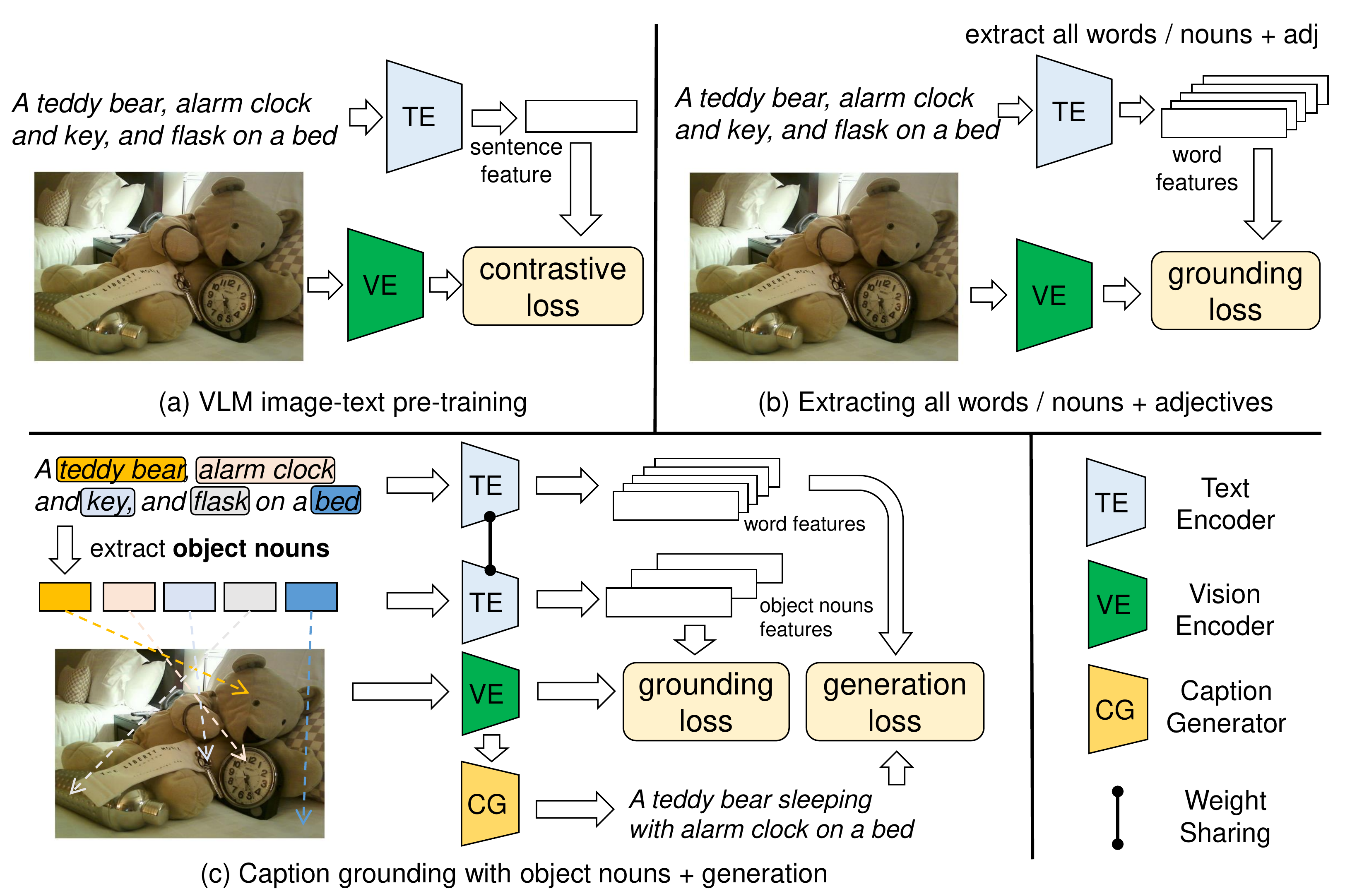}
	\caption{ (a) VLMs learn image-level visual-linguistic alignment using caption data. (b) Previous open vocabulary detection/segmentation methods extract all words~\cite{zareian2021open} or nouns + adjectives~\cite{OpenSeg} for caption grounding. (c) The proposed CGG extracts object nouns for a finer alignment between objects in the caption and visible entities in the image and then combines a caption generation loss to utilize the contextual knowledge in the caption fully.} 
	\label{fig:teaser_1}
\end{figure}

Instance Segmentation~\cite{coco_dataset} is a core vision task that goes beyond object detection~\cite{focal_loss,fpn,ren2015faster} via segmenting and classifying each object. Despite it continues to attract significant research effort~\cite{maskrcnn,tian2020conditional,yolact-iccv2019,wang2020solov2,zhou2022transvod,cheng2021mask2former,detr,chen2020blendmask,cheng2020panoptic,li2020panopticFCN,li2023tube,li2023transformer,zhang2023rethinking,zhang2021analogous}, current solutions mainly focus on a closed-set problem that assumes a pre-defined set of object categories~\cite{coco_dataset,OpenImages,lvis_data}. 
In practice, many applications need to detect and segment new categories. To save the need of annotating new object categories, zero-shot object detection/segmentation~\cite{zeroshotobjectdetection, zeroshotsegmentation} is proposed, where models are trained on base classes and equipped with the ability to segment new classes. However, the zero-shot setting performs poorly on novel classes, as high-level word embeddings cannot effectively encode fine-grained visual information.

\begin{figure*}[!t]
	\centering
	\includegraphics[width=1.0\linewidth]{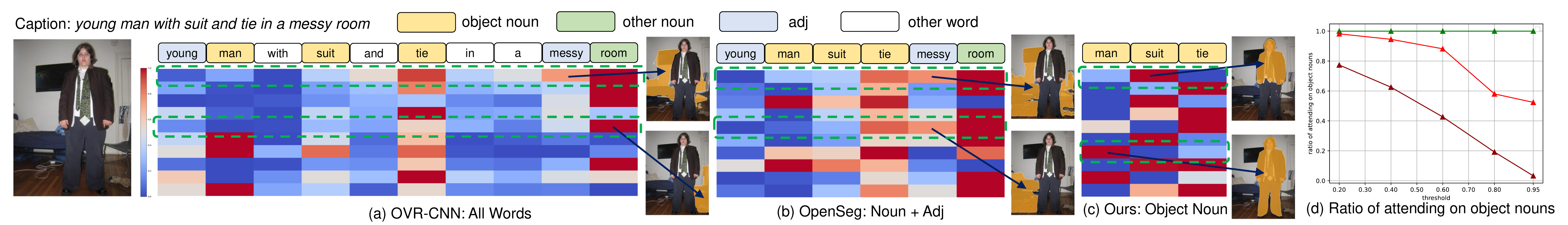}
	\caption{ A comparison analysis of caption grounding using different types of words. The color maps are normalized similarities between multi-modal embeddings and word features extracted by the language encoder. Both (a)~\cite{zareian2021open} and (b)~\cite{OpenSeg} suffer from the problem that invisible nouns (room in the example) are learned to be aligned by the multi-modal embeddings while using object nouns avoids the question. We adopt top-10 object queries according to their object scores. (d) We sample 2500 images from the COCO validation set and test the average rate of multi-modal embeddings attending on object nouns under different thresholds.}
	\label{fig:query_word_att}
\end{figure*}

To address this issue, recent work~\cite{zareian2021open} proposes an open vocabulary setting by pre-training a visual backbone on captioned images for learning rich visual features. 
With the success of pre-trained Vision Language Models (VLMs)~\cite{CLIP,ALIGN}, several approaches, \eg, ViLD \cite{ViLD}, propose effective methods to distill knowledge from VLMs into detectors or segmentation methods. Meanwhile, several works decouple the learning of open vocabulary classification and detection/segmentation into a two-stage pipeline~\cite{OpenSeg,ding2021decoupling}. Recently, state-of-the-art solutions~\cite{detic,promtdet,xpm,li2021grounded,zhang2022glipv2} for open vocabulary detection/segmentation try to adopt larger-scale dataset pre-training with the help of VLMs. For example, Detic~\cite{detic} adopts the ImageNet-21k~\cite{russakovsky2015imagenet} dataset to enlarge the detector in a weakly supervised manner, while PromptDet~\cite{promtdet} augments the detection dataset with image-caption pairs scraped from the Internet. Recent XPM~\cite{xpm} also pre-trains their model on caption datasets~\cite{sharma2018conceptual}. 
These approaches typically require a complex architecture design to leverage extra datasets~\cite{russakovsky2015imagenet,OpenImages}. Despite the performance improvement, these methods are not cost-effective in terms of data utilization. In this paper, we explore the use of caption data with more effective designs.

Caption-related vision tasks can be broadly divided into grounding and generation. The former~\cite{yu2018mattnet,VLT,liu2019learning,gonzalez2021panoptic} requires a model to align the text and corresponding region features, \eg, OVR-CNN~\cite{zareian2021open} and OpenSeg~\cite{OpenSeg} in Fig.~\ref{fig:teaser_1} (a) and (b).
However, these methods expose a core issue in that they adopt the grounding loss between words and mask regions, implicitly assuming each word (or noun) should correspond to a region in the image. As shown in Fig.~\ref{fig:query_word_att} (a) and (b), `messy' and `room' are forced to ground to meaningless masks. This motivates us to reformulate the ground loss by only focusing on object nouns as Fig.~\ref{fig:query_word_att} (c).
On the other hand, the latter~\cite{image-caption,image-caption1,image-caption2} learns a model that outputs a caption for a given imagery input. It naturally captures the auxiliary and surrounding information to generate context words, which is crucial to building the bridge between image and text. 
Given the above observation, we argue that caption generation can naturally complement the grounding loss for context capturing.

Therefore, we propose a unified framework based on Mask2Former~\cite{cheng2021mask2former} performing each task jointly to exploit the knowledge from caption data better. It contains a caption grounding loss and an extra caption decoder for the generation loss, as shown in \cref{fig:teaser_1} (c).
Motivated by the correlation analysis of object query and caption data (Sec.~\ref{sec:cgg_framework}), we first extract object nouns for grounding loss. 
In particular, we transform the object queries into multi-modal embeddings using a linear layer at the input stage. Then we adopt separated object nouns to ground each multi-modal embedding, providing us with the grounding loss. 
Since extracted object nouns miss the structure information of caption data, we append a caption generation loss in the output stage to recover language data. 
We add a lightweight Transformer decoder with multi-modal embeddings as inputs to generate captions. 
Experiments demonstrate that the two losses are well coupled and mutually affect novel class segmentation, with only \textbf{0.8\% GFlops} added during training. Our method drops the caption generation module for inference with no extra computation cost.
 
Our contributions can be summarized as follows:
\begin{itemize}
    \item We propose a joint Caption Grounding and Generation (CGG) framework for open vocabulary instance segmentation, which incorporates grounding with object nouns and caption generating.
    \item Experimental results demonstrate our method achieves a significant improvement of \textbf{6.8\% mAP} over previous XPM~\cite{xpm} on OVIS and \textbf{15\% PQ} improvements over previous method~\cite{DualOSPS} on OSPS.
\end{itemize}

\section{Related Work}

\noindent
\textbf{Zero-Shot Detection and Segmentation.} Collecting and annotating data on a large scale is laborious and expensive for detecting and segmenting in an extensive vocabulary. Zero-Shot Detection \cite{zeroshotobjectdetection} and Segmentation \cite{zeroshotsegmentation,D2Zero,PADing} aim to detect and segment novel categories that the annotations are not accessible during training. To address this problem, many studies align region features with fixed text embeddings \cite{fromedeep, zeroshotobjectdetection1, zeroshotobjectdetection2, zhang2021prototypical, zeroshotobjectdetection3}. However, due to the limited capacity of word embeddings and the emergence of large Vision-Language-Models (VLMs), recent studies~\cite{zareian2021open, ViLD, OV-DETR} have shifted towards the open vocabulary setting. 

\noindent
\textbf{Open Vocabulary Object Detection (OVOD).}  Recent studies~\cite{DetPro, zareian2021open, ViLD, detic, OV-DETR, wu2023open} focus on the open vocabulary setting, where models are trained 
 additionally on image-text pairs such as captions and text prompts.
For example, OVR-CNN \cite{zareian2021open} pre-trains on image-caption data to recognize novel objects, then fine-tunes the model for zero-shot detection. Recently, many works on image classification successfully expand their vocabulary sizes by pre-training on large-scale image-text pairs datasets. ViLD \cite{ViLD} distills the rich representation of pre-trained CLIP~\cite{CLIP} into the detector, while DetPro \cite{DetPro} adds a fine-grained automatic prompt learning. Meanwhile, several works extract pseudo-region annotations from the pre-trained VLMs and use them as additional training data for detectors. Detic \cite{detic} improves the performance of the novel classes with image classification datasets by supervising the max-size proposal with various image labels. These methods above share a common idea of enlarging the capacity of training data to find rare classes, but they require more computation/annotation costs and complex pipelines. In contrast, we design a way to discover novel classes from caption data in one unified framework \textbf{\textit{without}} pre-training on extra datasets nor distilling knowledge from pre-trained VLMs.
 
\noindent
\textbf{Open Vocabulary Segmentation (OVS).} Beyond OVOD, OVS further requires the model to segment the novel classes. Current solutions for OVS usually decouple mask generation and mask classification into two different steps. The former generates mask regions, while the latter performs classification with pre-trained VLMs~\cite{OpenSeg, LSeg}. DenseCLIP~\cite{DenseCLIP} proposes a similar pipeline to OVOD by distilling CLIP knowledge through generating pseudo mask labels. 
Our method proposes an end-to-end pipeline that jointly performs caption learning (grounding/generation) and segmentation learning. XPM~\cite{xpm} proposes a cross-modal pseudo-labeling framework by aligning word features in captions with visual features in images.

\noindent
\textbf{Image Captioning.} This task requires the model to generate captions that describe the content of images~\cite{image-caption}. State-of-the-art methods use multi-modal attention designs, treating the task as a multi-modal translation problem~\cite{image-caption1, image-caption2, image-caption3}. Our focus in this work is not on designing a new captioning model, but on exploring image captioning as a sub-task for open vocabulary learning to enhance the novel class discovery ability. Using caption generation as an auxiliary loss is also adopted in vision language pre-training~\cite{cho2021unifying,luo2020univl}. However, to our knowledge, this is the \textit{first study} exploring caption generation for OVS.

\begin{figure*}[t!]
	\centering
	\includegraphics[width=0.90\linewidth]{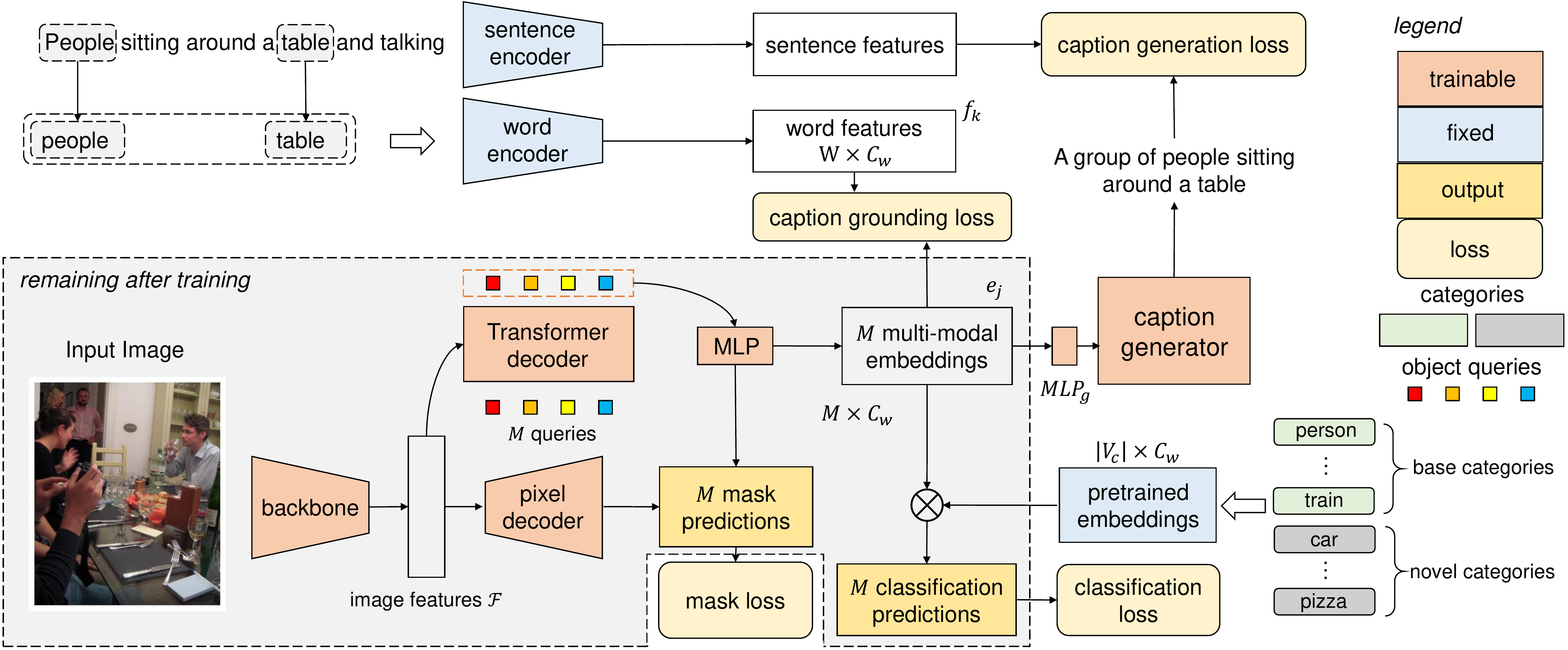}
	\caption{ The illustration of \textbf{CGG} framework. The input image $I$ is first provided to Mask2Former. The output of the Transformer decoder is then fed into an MLP, which generates $M$ mask predictions together with the output of the pixel decoder in one hand. On the other hand, the object queries are transferred into $M$ multi-modal embeddings, denoted as $\{e_j|j\in\{1,2,\cdots,M\}\}$. The similarities of these embeddings with class embeddings are then computed to produce classification predictions. $\{e_j\}$ are also involved with grounding loss and generation loss with text features extracted by word and sentence encoder.}
\label{fig:methods}
\end{figure*}

\section{Methodology}
%

In this section, we first review the background of OVIS and the baseline as preliminary. Then, we carry out the analysis on the correlation of caption data and query-based segmenter. Next, we present our Caption Grounding and Generation framework, which aims to exploit caption data via joint caption grounding and generation.

\subsection{Preliminary}

\noindent
\textbf{Problem Setting.} 
We first describe the open-vocabulary problem setting.
Let $\mathcal{D}_B=\{(I_m, M_m)\}_{m=1}^{N_B}$ be the set of training images and instance annotations for a limited set of base classes $\mathcal{V}_B$. Among these images, there are also novel classes $\mathcal{V}_N$, whose annotations cannot be accessed during the training. 
%
%
Each image $I_m$ is associated with a set of ground-truth (GT) annotations $M_m$, which comprises instance masks and their corresponding object classes. To detect and segment novel classes, following previous works~\cite{zareian2021open}, we leverage additional image-level annotations, i.e., image captions. Let $\mathcal{D}_C=\{(I_c, C_c)\}_{c=1}^{N_C}$ be another set of training images with image caption annotations. Each image $I_c$ is annotated with a caption $C_c$. Compared to pixel-level annotations, captions are easier to collect, and its vocabulary $\mathcal{V}_C$ is much larger than base classes, i.e., $|\mathcal{V}_C| \gg |\mathcal{V}_B|$.
Therefore, exploiting the additional information from the image caption dataset would be beneficial.
OVIS aims to train a model to segment both base classes $\mathcal{V}_B$ and novel classes $\mathcal{V}_N$. Following previous methods~\cite{zareian2021open, xpm, OpenSeg}, our model uses high-level semantic embeddings from a pre-trained text Transformer (BERT~\cite{BERT}) as the weights of the linear classifier. We focus on distilling knowledge in the captions to the target classes via representation similarities.
In the following sections, we will neglect the image index for simplicity.

\noindent
\textbf{Baseline Method.} We adopt the recent Mask2Former~\cite{cheng2021mask2former} model as our baseline since the query-based Transformer architecture can be readily extended into multi-modal training with captions. 
Given an image $I$, during the inference, Mask2Former directly outputs a set of $M$ object queries $\Q=\{q_j|j=1,..,M\}$, where each object query $q_j$ represents one entity.
Then, two different Multiple Layer Perceptrons (MLPs) project the queries into two embeddings for mask classification and prediction. During training, a bipartite matching algorithm matches each object query to the ground truth mask, following~\cite{cheng2021mask2former}. The loss function is $L_{mask} = \lambda_{cls}L_{cls} + \lambda_{ce}L_{ce} + \lambda_{dice}L_{dice}$, where $L_{cls}$ is the Cross-Entropy (CE) loss for mask classification, and $L_{ce}$ and $L_{dice}$ are the Cross-Entropy (CE) loss and Dice loss~\cite{dice_loss} for segmentation, respectively. In particular, following~\cite{zareian2021open}, we use pre-trained embeddings to replace the learnable classifier for training and inference, as shown in Fig.~\ref{fig:methods}. However, the original Mask2Former can only detect and segment closed-set objects and cannot handle the novel classes. Our method extends it to perform open-vocabulary segmentation in a new framework.

\subsection{CGG Framework for OVS}
\label{sec:cgg_framework}

\noindent
\textbf{Overview.} Fig.~\ref{fig:methods} presents the overall pipeline of the CGG framework. Following~\cite{zareian2021open}, we set the pre-trained text embeddings as the weights of the linear classifier. Then we add two losses: the caption grounding loss and the caption generation loss. A caption generator is appended at the end of the output queries, producing the image caption. During training, we adopt a pre-trained sentence encoder and word encoder to encode both captions and object nouns extracted from captions into sentence and word features. The former is used for caption generation, while the latter is for caption grounding. We discard all newly-introduced modules during inference and perform a lightweight inference procedure.

\noindent
\textbf{Analysis on Grounding Target with Object Query.} 
Previous works like OVR-CNN~\cite{zareian2021open} pre-train their models with caption data.  However, there are two potential issues with the previous design. Firstly, training caption and segmentation separately cannot fully explore caption data and detection/segmentation annotations. The training of the segmenter is isolated, so the connection between the two models is broken. Secondly, there is a weakened region-word alignment in the traditional grounding process by calculating similarities between multi-modal embeddings and \textbf{all} words in caption data, because object-unrelated words may encounter the vision-language implicit matching. 

For the first problem, we adopt a query-based detector~\cite{cheng2021mask2former} for end-to-end co-training. For the second problem, we argue that object nouns in caption data should be well aligned with query features in a more fine-grained manner since \textit{the novel class categories are always nouns}. In Fig.~\ref{fig:query_word_att} (a)-(c), we visualize the attention map of multi-modal embeddings and extracted word features, where we find several background items like rooms also have a high similarity with multi-modal embeddings, which brings the noise in supervision. In Fig.~\ref{fig:query_word_att} (d), we perform a statistical analysis on the ratio of attended nouns, finding a significant drop with the increase of thresholds. Since object queries with higher scores always play as the output of instance segmentation, we argue this may hurt the performance of final segmentation results. Combining the above analysis and findings, we propose adopting object nouns as grounding targets.

\noindent
\textbf{Caption Grounding with Object Nouns.} 
For the image-caption pair $(I,C)$, we first extract object nouns from the caption $C$ and feed it to the word encoder. Here, we neglect the image index for simplicity.
We get word features $\{f_k |\ k \in \{1, 2, ..., K\}\}$, where $K$ is the number of tokens from object nouns. For the image input $I$, we adopt an MLP layer to project the output of the Transformer decoder to a set of multi-modal embeddings $\{e_j |\ j \in \{1, 2, ..., M\}\}$, where $M$ is the number of object queries in Mask2Former. 

\begin{figure}[!t]
	\centering
	\includegraphics[width=1\linewidth]{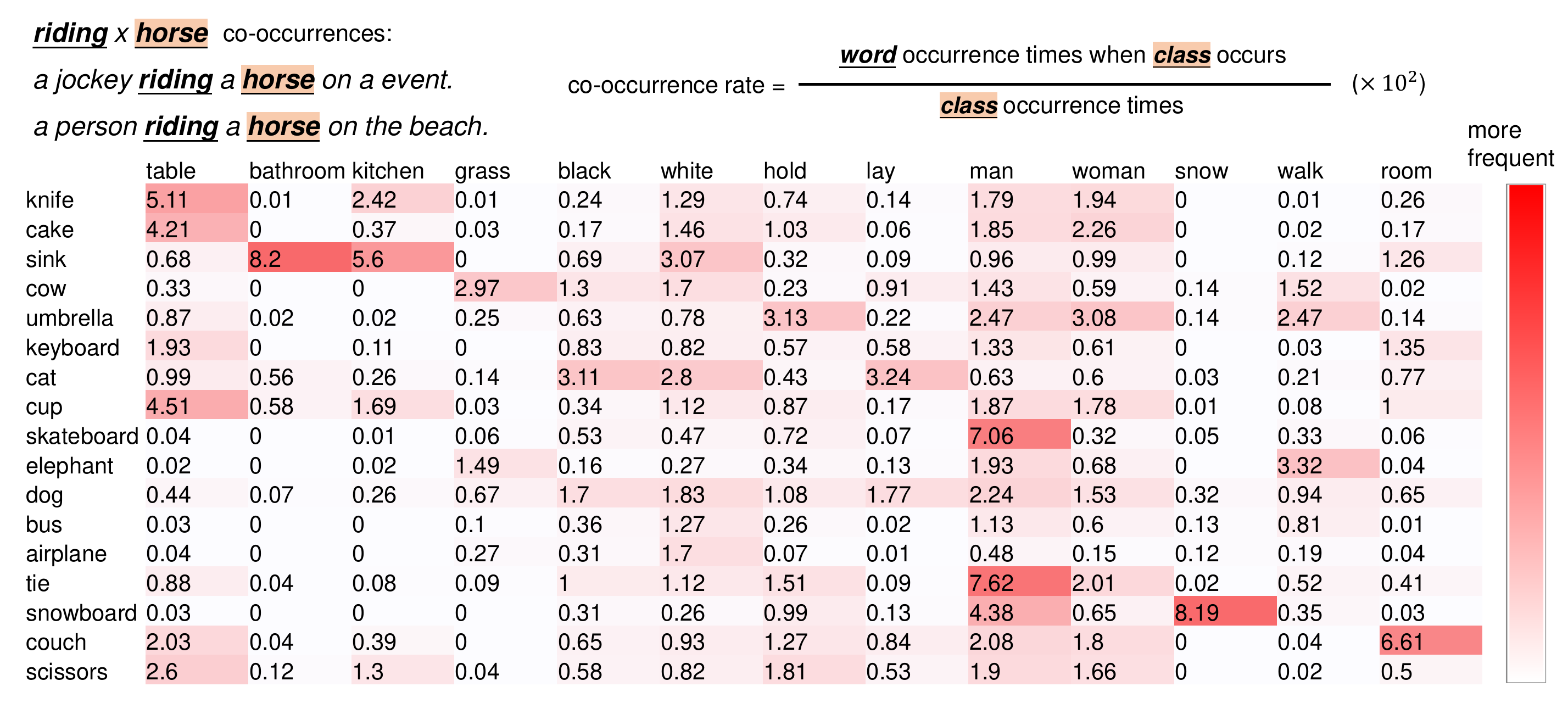}
	\caption{ We often observe certain pairs of words co-occurrence while others do not. We calculate the co-occurrence matrix between novel classes and frequent words in the caption. Different classes have various distributions on co-occurrence words.}
	\label{fig:co-occurrence}
\end{figure}
The similarity between image $I$ and caption $C$ is calculated as:
\begin{equation}
\begin{aligned}
\scriptsize
S^C(I,C) &= \frac{1}{M}\sum^{M}_{j=1}\sum^{K}_{k=1}a^I_{j,k}\langle e_j,f_k \rangle,
\end{aligned}
\end{equation}
where $\langle \cdot, \cdot \rangle$ is a dot production operation.  $S^C(I,C)$ is normalized along the text dimension. 
$a^C_{j,k} = \exp \langle e_j, f_k\rangle / \sum^{K}_{l=1} \exp \langle e_j,f_l \rangle $
is the normalization term. Similarly, we can also get $S^I(I,C')$ by normalizing along the image dimension.

During training, the similarities between matching image-caption pairs should be maximized. 
For a mini-batch of image-caption pairs input $(\mathbf{I}, \mathbf{C})$, the objective function is:
\begin{equation}
L^{CC}_{gro}(I) = -\log \frac{\exp S^C(I,C)}{\sum_{C' \in \mathbf{C}}\exp S^C(I,C')},
\end{equation}
and by normalizing along the image dimension, there is
\begin{equation}
L^{CI}_{gro}(I) = -\log \frac{\exp S^C(I,C)}{\sum_{I' \in \mathbf{I}}\exp S^C(I',C)}.
\end{equation}
Similarly, we can get $L^{IC}_{gro}(I)$ and $L^{II}_{gro}(C)$ using $S^I(I, C)$.
The final grounding loss for the batch is the summation of the four losses,
\begin{equation}
\left.
\begin{aligned}
L_{gro} = &\frac{1}{|\mathbf{I}|}\sum_{I}(L^{CC}_{gro}(I) + L^{IC}_{gro}(I))\ + \\
&\frac{1}{|\mathbf{C}|}\sum_{C}(L^{CI}_{gro}(C) + L^{II}_{gro}(C)).
\end{aligned}
\right.
\end{equation}

\begin{figure}[!t]
	\centering
	\includegraphics[width=1\linewidth]{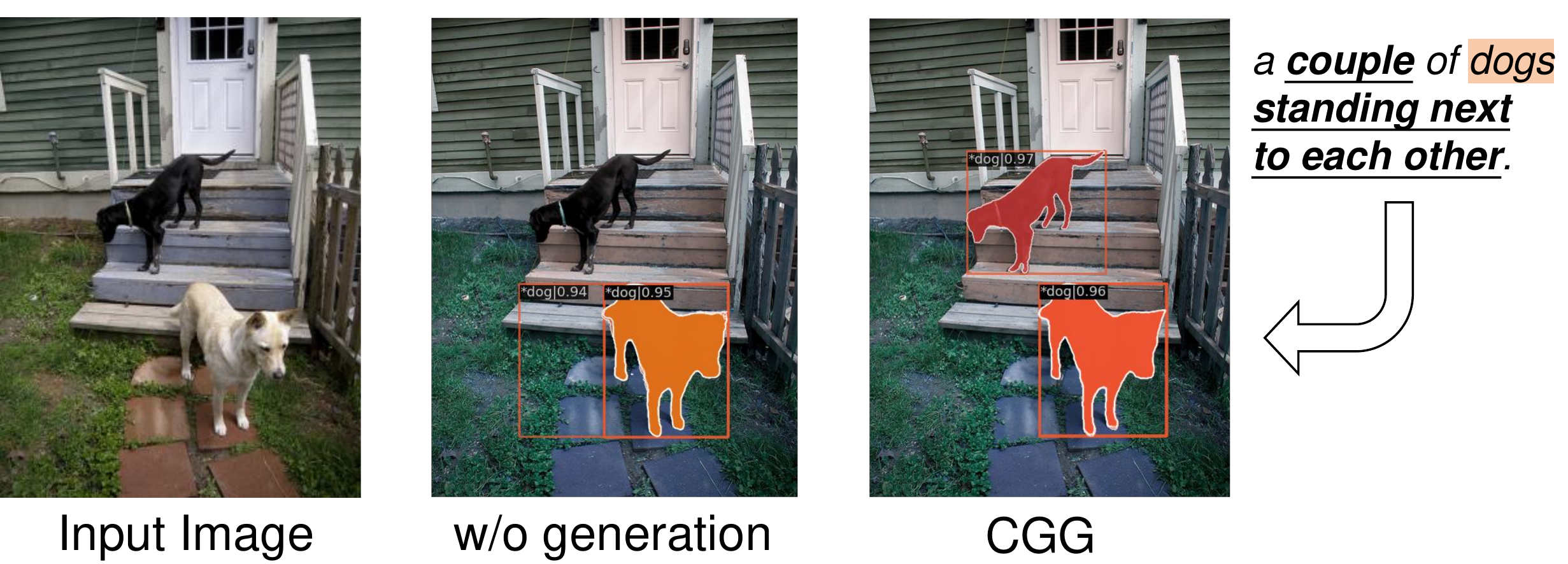}
	\caption{ The effectiveness of caption generation. The generated caption depicts rich information beyond object nouns.}
	\label{fig:generation_visualization}
\end{figure}

\noindent
\textbf{Grounding Object Nouns Misses the Structure Information of Caption Data.} Despite grounding nouns forces to push the nouns embeddings and object queries closer, the structure information, including the relation of different objects is missing. As shown in Fig.~\ref{fig:co-occurrence}, we perform co-occurrence relationship analysis on object nouns and find that novel classes have various distributions on co-occurrence words, which may help identify novel objects. This means only adopting grounding loss misses the relationship between these words. To fill such a gap, we argue that caption data can also be employed as a generative supervision signal for a more fine-grained multi-modal understanding.
The key insight is that we force the model to predict the occurring instances and their relationships in the image to identify novel classes. Unlike grounding loss that aims to push nouns and query embeddings as close as possible, generative loss decodes the visual features into the semantic embeddings, which are complementary to grounding loss. As shown in Fig. \ref{fig:generation_visualization}, the caption generation module can help the model learn the specific status and relationships of objects in the scene.

\noindent
\textbf{End-to-End Caption Generation Loss.}
Specifically, since the multi-modal embeddings encode the region-wise information, we transform these embeddings using a linear layer ($MLP_{g}$ in Fig.~\ref{fig:methods}) to fit the feature dimension of the caption generator, then we directly take the transformed embeddings as the input of the lightweight caption generator, which includes a stack of Transformer decoder layers. We adopt a Cross Entropy Loss on the predicted distribution of text vocabularies. It is the commonly used objective function in the research field of caption generation. 
\begin{equation}
L_{gen} = -\sum^{N_s}_{t=1}\log (p_\theta (\hat{w}_t|w_1, ..., w_{t-1})),
\end{equation}
where $p_\theta (\hat{w}_t | w_1,\dots,w_{t-1})$ is the predicted probability of $t$-th right word over the whole vocabulary, $\theta$ denotes the parameters of the generation network. Hence, this loss function enforces the predicted sentence to be consistent with the input caption $C$, making the multi-modal embeddings $\{e_j|j\in\{1,2,\cdots,M\}\}$ capable of representing various words and their potential relations in the image.

\noindent
\textbf{Overall Loss Design.} The overall training loss contains four items, i.e., the classification loss $L_{cls}$, the segmentation loss $L_{mask}$, the caption grounding loss $L_{gro}$, and the caption generation loss $L_{gen}$. Following the previous method \cite{zareian2021open}, the classification loss is selected as the Cross-Entropy Loss that takes the dot product of multi-modal embeddings $e^M_i$ and base class embeddings as its logit inputs. The final loss function $L$ is the weighted summation of the four losses: $\label{equ:over_all_loss} L = \lambda_{cls}L_{cls} + \lambda_{mask}L_{mask} + \lambda_{gro}L_{gro} + \lambda_{gen}L_{gen}$. We follow the default setting in the MMDetection framework, where the weights are set to 2.0, 5.0, 2.0, and 2.0 in all our experiments.


\noindent
\textbf{Training and Inference.} Compared to the baseline model, CGG only introduces extra losses and a caption generation head during the training. Following previous works \cite{OpenSeg, zareian2021open, xpm}, we first pre-train our framework using \textit{only} base data annotations in a class-agnostic manner. The goal of pretraining is to encode instance-wised information into object queries. Then we load the pre-trained model for joint training with caption data. During the inference, following~\cite{zareian2021open}, we use the pre-trained embeddings of all classes to perform open vocabulary segmentation via dot product, including base classes and novel classes.
\begin{table}[!t]
   \centering
    \caption{ Results on Open Vocabulary Instance Segmentation.}
   \scalebox{0.9}{
   \setlength{\tabcolsep}{2.5mm}{\begin{tabular}{c|c|c|c|c|c}
      \toprule[0.15em]
      \multirow{2}{*}{Method} & \multicolumn{2}{c|}{Constrained}  & \multicolumn{3}{c}{Generalized} \\
        \cline{2-6}
        & Base & Novel & Base & Novel & All \\
        \hline
        OVR \cite{zareian2021open} & 42.0 & 20.9 & 41.6 & 17.1 & 35.2 \\
        SB \cite{zeroshotobjectdetection1} & 41.6 & 20.8 & 41.0 & 16.0 & 34.5 \\
        BA-RPN \cite{BA-RPN} & 41.8 & 20.1 & 41.3 & 15.4 & 34.5 \\
        XPM \cite{xpm} & 42.4 & 24.0 & 41.5 & 21.6 & 36.3 \\
        \hline
        CGG (Ours) & \textbf{46.8} & \textbf{29.5} & \textbf{46.0} & \textbf{28.4} & \textbf{41.4} \\
      \bottomrule[0.10em]
   \end{tabular}}}
   \label{tab:result_OVIS}
\end{table}

\begin{table}[t]
  \centering
    \caption{ Results on COCO Open Vocabulary Object Detection (OVOD). IN-21K indicates ImageNet-21K~\cite{imagenet}. CC indicates Conceptual Captions \cite{conceptual-captions}}
  \scalebox{0.83}{
  \begin{tabular}{l c c c c c c }
    \toprule[0.2em]
    {Method} & Epochs & Extra Data & AP50$_{\text{novel}}^{\text{box}}$ & AP50$_{\text{all}}^{\text{box}}$  \\
   \hline
   DLWL \cite{DLWL} & 96 & YFCC100M & 19.6 & 42.9 \\
   Cap2Det \cite{cap2det} & 8.5 & None & 20.3 & 20.1 \\
   OVR-CNN \cite{zareian2021open} & 12 & None & 22.8 & 39.9 \\
   Detic \cite{detic} & 96 & IN-21K \& CC & 24.1 & 44.7 \\
   PromptDet \cite{promtdet} & 24 & LAION-novel & 26.6 & 50.6 \\
    \hline
    CGG (Ours) & 12 & None & \textbf{29.3} & 42.8 \\ 
    \bottomrule[0.10em]
  \end{tabular} 
  }
  \label{tab:result_OVOD}
\end{table}

\section{Experiments}

\subsection{Experimental Setup}
\label{exp:set_up}


\noindent
\textbf{Dataset Settings.} We conduct experiments on COCO dataset \cite{coco_dataset} for OVIS and OSPS. 
For OVIS, following previous works \cite{zareian2021open, xpm}, we split 48 base classes with annotations and 17 target classes without annotations. For captioned images, we use the entire COCO-captions training set with 118,287 images and five captions per image. 
Unlike previous works~\cite{detic, promtdet, DLWL} that adopt extra caption datasets, like Conceptual Captions~\cite{conceptual-captions} for pre-training, we do \textbf{not} use extra caption or detection datasets. We follow the origin OVR-CNN~\cite{zareian2021open} setting by only exploring a limited caption dataset within COCO. 
For OSPS~\cite{eopsn}, we follow the previous works \cite{eopsn, DualOSPS}, splitting part of thing classes into unknown classes. 
We obtain three different splits by varying the numbers of unknown classes ($K\%$ ratios, 5\%, 10\%, 20\%). 

\begin{table}[!t]
   \centering
      \caption{ We compare our method CGG with previous methods EOPSN \cite{eopsn} and Dual \cite{DualOSPS} on Open Set Panoptic Segmentation (OSPS). Unlike EOPSN and Dual that group all unknown things into one class without identifying them, CGG performs Open Vocabulary Panoptic Segmentation and assigns a specific category to each unknown thing. We show the mean PQ and SQ for all unknown categories and indicate the scores averaged from each unknown class with ``*''.}
   \scalebox{0.75}{
   \setlength{\tabcolsep}{2.5mm}{\begin{tabular}{c|c|c|c|c|c|c|c}
      \toprule[0.15em]
      \multirow{2}{*}{Method} & \multirow{2}{*}{$K(\%)$} &\multicolumn{4}{c|}{Known}  & \multicolumn{2}{c}{Unknown} \\
        \cline{3-8}
         & & $\text{PQ}^{\text{Th}}$ & $\text{SQ}^{\text{Th}}$& $\text{PQ}^{\text{St}}$ & $\text{SQ}^{\text{St}}$ & $\text{PQ}^{\text{Th}}$ & $\text{SQ}^{\text{Th}}$\\
        \hline
        EOPSN \cite{eopsn} & \multirow{3}{*}{5} & 44.8 & 80.5 & 28.3 & 73.1 & 23.1 & 74.7  \\
        Dual \cite{DualOSPS} & & 45.1 & 80.9 & 28.1 & 73.1 & 30.2 & 80.0 \\
        CGG (Ours) &  & \textbf{50.2} & \textbf{83.1} & \textbf{34.3} & \textbf{81.5} & \textbf{45.0*} & \textbf{85.2*}  \\
        \hline
        EOPSN \cite{eopsn} & \multirow{3}{*}{10} & 44.5 & 80.6 & 28.4 & 71.8 & 17.9 & 76.8 \\
        Dual \cite{DualOSPS}& & 45.0 & 80.7 & 27.8 & 72.2 & 24.5 & 79.9 \\
        CGG (Ours) & & \textbf{49.2} & \textbf{82.8} & \textbf{34.6} & \textbf{81.2} & \textbf{41.6*} & \textbf{82.6*} \\
        \hline
        EOPSN \cite{eopsn} & \multirow{3}{*}{20} & 45.0 & 80.3 & 28.2 & 71.2 & 11.3 & 73.8  \\
        Dual \cite{DualOSPS}& & 45.0 & 80.6 & 27.6 & 70.1 & 21.4 & \textbf{79.1} \\
        CGG (Ours) & & \textbf{48.4} & \textbf{82.3} & \textbf{34.4} & \textbf{81.1} & \textbf{36.5*} & 78.0* \\
      \bottomrule[0.10em]
   \end{tabular}}}
   \label{tab:result_OSPS}
\end{table}

\noindent
\textbf{Metric.} For OVIS, we report the mask-based mean Average Precision (mAP) at intersection-over-union (IoU) of 0.5. Following previous works~\cite{zareian2021open, xpm}, we evaluate the model performance on base and target classes in two settings: constrained setting, where the model is only tested on images that belong to either base or target classes; generalized setting, where the model is tested on both base and target classes. The latter is more challenging, as it requires the model to avoid class bias from base classes. We also report open vocabulary detection with box-based mAP. For OSPS setting, we use panoptic segmentation metrics, including Panoptic Quality (PQ) and Segmentation Quality (SQ). We report known classes and unknown classes separately for reference. More details about the data preparation can be found in the appendix.

\begin{table*}[!t]
    \footnotesize
	\centering
	\caption{ Ablation studies and analysis on COCO OVIS.}
    \subfloat[The Effectiveness of Each Components.]{
    \label{tab:ablation_a}
	    \begin{tabularx}{0.33\textwidth}{c c c c c c c} 
		\toprule[0.15em]
    	baseline & Gro. & Gen. & Base & Novel  \\
        \midrule[0.15em]
        Class Emb.    &  &  & 48.6 & 0.2 \\
        w. Gro. & \checkmark &  & 49.1 & 22.2 \\
        w. Gen. & & \checkmark & 49.4 & 0.3  \\
        %
        
        %
        
    	Both (CGG) & \checkmark & \checkmark & 48.0 & \textbf{28.4} \\
        \bottomrule[0.1em]
	    \end{tabularx}
    } \hfill
    \subfloat[Training Pipeline Comparison]{
     \label{tab:ablation_b}
	    \begin{tabularx}{0.30\textwidth}{c c c c} 
		        				\toprule[0.15em]
    		 Settings  & Base & Novel & All \\
    		 \midrule[0.15em]
    		     emb-segm & 49.2 & 20.3 & 41.6 \\
    	       segm-emb-segm & \textbf{50.2} & 24.3 & \textbf{43.4} \\
    	       segm-emb (CGG) & 46.0 & \textbf{28.4} & 41.4  \\
        	\bottomrule[0.1em]
	    \end{tabularx}
    }
    \hfill
    \subfloat[Nouns Extraction in Caption Grounding]{
    \label{tab:ablation_c}
		\begin{tabularx}{0.33\textwidth}{c c c c} 
			\toprule[0.15em]
			Method & Base & Novel & All  \\
			\midrule[0.15em]
            All Words & 44.7 & 7.6 & 35.0 \\
            Nouns + Adj & 46.2 & 16.2 & 39.2 \\
            Object Nouns + Adj & 45.6 & 27.2 & 40.2 \\
            Object Nouns & 46.0 & \textbf{28.4} & 41.4 \\
			\bottomrule[0.1em]
		\end{tabularx}
    } \hfill
    \subfloat[Caption Generator Design]{
     \label{tab:ablation_d}
	    \begin{tabularx}{0.28\textwidth}{c c c c} 
		        				\toprule[0.15em]
    		 \#layers & Base & Novel & All  \\
    		\midrule[0.15em]
            2 & 46.7 & 23.4 & 40.6 \\
    	    4 & 46.0 & \textbf{28.4} & 41.4 \\
            6 & 48.2 & 26.9 & 42.6  \\
        	\bottomrule[0.1em]
	    \end{tabularx}
    } \hfill
    \subfloat[Effect of Class-Agnostic Pretraining]{
     \label{tab:ablation_e}
	    \begin{tabularx}{0.32\textwidth}{c  c  c c} 
		        				\toprule[0.15em]
    		 Settings & Base & Novel & All \\
    		\midrule[0.15em]
    	   No class-agnostic  &  46.2 & 22.7 & 40.0 \\
              Freeze class-agnostic & 47.6 & 26.4 & 42.1 \\
              CGG & 46.0 & \textbf{28.4} & 41.4 \\
        	\bottomrule[0.1em]
	    \end{tabularx}
    } \hfill
    \subfloat[GFlops and Parameters]{
    \label{tab:ablation_f}
		\begin{tabularx}{0.30\textwidth}{c c c} 
			\toprule[0.15em]
			Schedule & Parameters & GFLOPs \\
			\midrule[0.15em]
            baseline & 35.65M & 227.48 \\
            Ours: Inference & 35.65M & 227.48 \\
            Ours: Training & 81.19M  & 229.33 \\
			\bottomrule[0.1em]
		\end{tabularx}
    } \hfill
\end{table*}

\noindent
\textbf{Implementation Details.} We implement our models in PyTorch~\cite{pytorch_paper} with MMDetection framework~\cite{chen2019mmdetection}. We use 8 GPUs for distributed training. Each mini-batch has two images per GPU. The optimizer is AdamW~\cite{ADAMW} with a weight decay of 0.0001. We adopt full image size for a random crop in the pre-training and training process following~\cite{cheng2021mask2former}. We use BERT embeddings~\cite{BERT} for the classification head, word encoder, and sentence encoder. We use an LVIS class name parser to extract object nouns from caption data. For OVIS, we keep the top 100 queries as the model outputs. For OSPS, we follow previous works \cite{eopsn, DualOSPS}, which put thing mask predictions first, then fill the remaining background with stuff mask predictions. We use the ResNet-50 backbone for all experiments for a fair comparison.

\subsection{Main Results}
\label{sec:exp_res}

\begin{table}[!t]
   \centering
    \caption{ Ablation on the ability of caption generation.}
   \scalebox{0.70}{
   \setlength{\tabcolsep}{1.5mm}{\begin{tabular}{c|c|c|c|c|c|c}
      \toprule[0.15em]
      \#layers & BLUE-1 $\uparrow$ & BLUE-2 $\uparrow$ & BLUE-3 $\uparrow$ & BLUE-4 $\uparrow$ & CIDEr $\uparrow$ & ROUGE $\uparrow$ \\
        \hline
        2 & 0.473 & 0.311 & 0.206 & 0.141 & 0.307 & 0.360 \\
        4 & 0.418 & 0.258 & 0.166 & 0.111 & 0.239 & 0.320 \\
        6 & 0.387 & 0.226 & 0.138 & 0.088 & 0.171 & 0.289 \\
      \bottomrule[0.10em]
   \end{tabular}}}
   \label{tab:ablation-generation-layer}
\end{table}

\noindent
\textbf{Results on OVIS.}
We first compare CGG and other methods for the OVIS task. Tab. \ref{tab:result_OVIS} shows that our model outperforms XPM, the best baseline, by 5.5\% mAP in the constrained setting and 6.8\% mAP in the generalized setting where both base and novel categories are employed as input. 
The generalized setting is more challenging because the model must distinguish novel categories from base categories, where the training data bias is for base categories. 
CGG has improved more in generalized than constrained settings, demonstrating its effectiveness in identifying and distinguishing novel classes from base classes.

\noindent
\textbf{Results on OVOD.}
We further evaluate our model on the Open Vocabulary Object Detection task, which requires matching ground truth with predicted bounding boxes at test time. 
Tab. \ref{tab:result_OVOD} shows that CGG outperforms several previous works~\cite{promtdet, detic} on novel classes in terms of AP50 score while using only COCO-Captions as the image-text data source and a shorter training schedule. 
Previous methods such as PromptDet \cite{promtdet} and Detic \cite{detic} rely on large-scale image-text datasets, which incur a longer training time and higher computational cost. 
However, CGG performs worse on all classes cases: AP$50_{all}^{box}$. 
It may be due to the limited exposure to base classes and shorter training schedules compared with other methods.

\begin{table}[!t]
   \centering
    \caption{ Comparison between only training caption generation and joint training with segmentation. ``only-gen'' means the model is trained purely with caption generation supervision.}
   \scalebox{0.70}{
   \setlength{\tabcolsep}{1.4mm}{\begin{tabular}{c|c|c|c|c|c|c}
      \toprule[0.15em]
      Method & BLUE-1 $\uparrow$ & BLUE-2 $\uparrow$ & BLUE-3 $\uparrow$ & BLUE-4 $\uparrow$ & CIDEr $\uparrow$ & ROUGE $\uparrow$ \\
        \hline
        only gen. & 0.394 & 0.237 & 0.150 &0.100 & 0.177 & 0.305 \\
        CGG & \textbf{0.418} & \textbf{0.258} & \textbf{0.166} & \textbf{0.111} & \textbf{0.239} & \textbf{0.320} \\
      \bottomrule[0.10em]
   \end{tabular}}}
   \label{tab:ablation-generation-compare}
\end{table}

\noindent
\textbf{Results on OSPS.}
We test CGG on the Open Set Panoptic Segmentation task by expanding the base classes from base thing classes to including stuff classes without changing the training pipeline of CGG. 
Tab. \ref{tab:result_OSPS} indicates that our model performs better than previous methods EOPSN~\cite{eopsn} and Dual~\cite{DualOSPS} by 14.9\% PQ on unknown things in 20\% unknown things setting, and 16.9\%, 14.8\% in 10\% and 5\% settings, respectively. 
Compared with the standard Open Set Panoptic Segmentation task, CGG classifies each unknown class and still outperforms previous methods.

\subsection{Ablation Study and Analysis}
\label{sec:exp_ablation}
To evaluate the effectiveness of each component of our model, we conduct ablation studies on the COCO 48/17 split \cite{zareian2021open} using mAP as the metric.

\begin{figure*}[!t]
	\centering
	\includegraphics[width=0.95\linewidth]{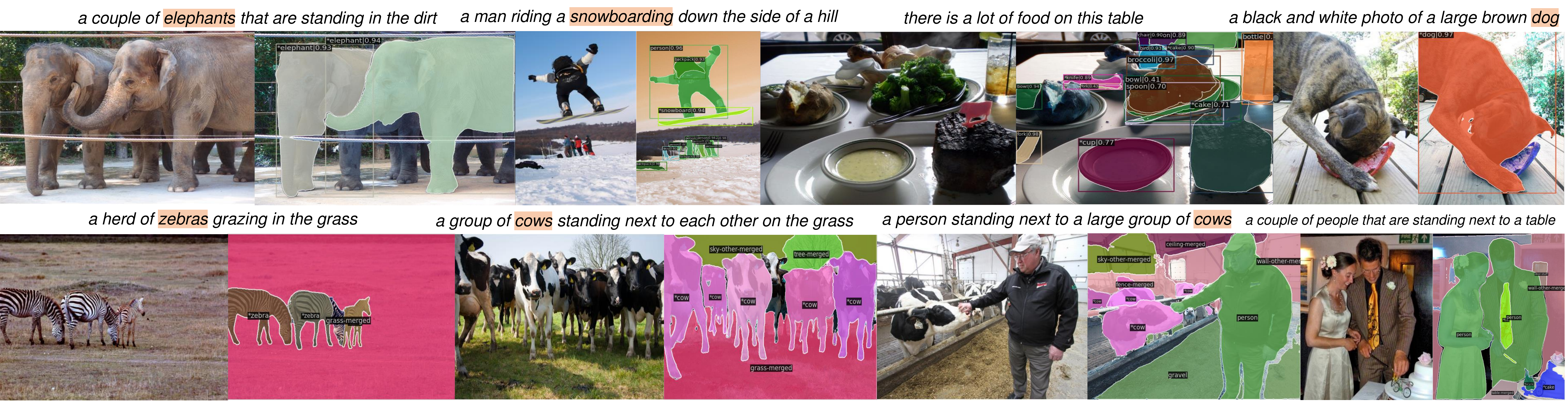}
	\caption{ Visualization results on Instance Segmentation (Top) and Panoptic Segmentation (Bottom). The categories with ``*'' are novel. We also generate captions for each image-prediction pair and highlight the novel categories in the captions, if any.}
	\label{fig:visual_results.}
\end{figure*}

\noindent
\textbf{Effectiveness of Modules.}
We first verify the effectiveness of each proposed module in CGG.
Tab. \ref{tab:ablation_a} shows that the baseline Class Emb., which maps class labels to text embeddings, achieves a low AP score of $0.2$ for the novel class. 
By contrast, adding Caption Grounding boosts the Novel AP to $22.2$, demonstrating the importance of Caption Grounding for aligning multi-modal embeddings to object nouns. 
The final score reaches $28.4$. 
This improvement comes from Caption Generation, which supervises object nouns and other meaningful words. 
Without Caption Grounding, Caption Generation alone performs poorly with $0.3$ AP. This observation demonstrates that Caption Grounding is the crucial module.

\noindent
\textbf{Training Pipeline.}
We compare different training pipelines for our model. 
Previous methods like OVR-CNN \cite{zareian2021open} use an ``emb-segm'' pipeline, which trains with captions first and then fine-tunes the segmentor. 
In contrast, we adopt a ``segm-emb'' pipeline, which pre-trains a class-agnostic segmentor and then trains the multi-modal embeddings $e^M_i$ on image-text data. Tab. \ref{tab:ablation_b} compares these pipelines over CGG. 
We also include ``segm-emb-segm'' as a candidate. The results indicate that although ``segm-emb'' performs worse than others for base classes, it achieves much higher scores for novel classes. 
Training the segmentor in the last stage causes overfitting on the base classes and reduces recall for novel classes.

\noindent
\textbf{Grounding Nouns Extraction.}
We investigate different word selection strategies for CGG as discussed in Sec.~\ref{sec:cgg_framework}. 
Instead of extracting only object nouns from the sentences, we extract all words~\cite{zareian2021open}, nouns + adj\cite{OpenSeg}, and object nouns + adj. Tab. \ref{tab:ablation_c} shows the results of these strategies. 
Extracting all words leads to a $20.8$ drop in novel class AP, and extracting nouns + adj leads to a $12.2$ drop in novel class AP. 
When extracting object nouns + adj, the performance is close to ours. 
These results indicate that selecting suitable words is crucial for our model's performance, where we find object nouns perform the best.

\begin{figure}[!t]
	\centering
	\includegraphics[width=1.0\linewidth]{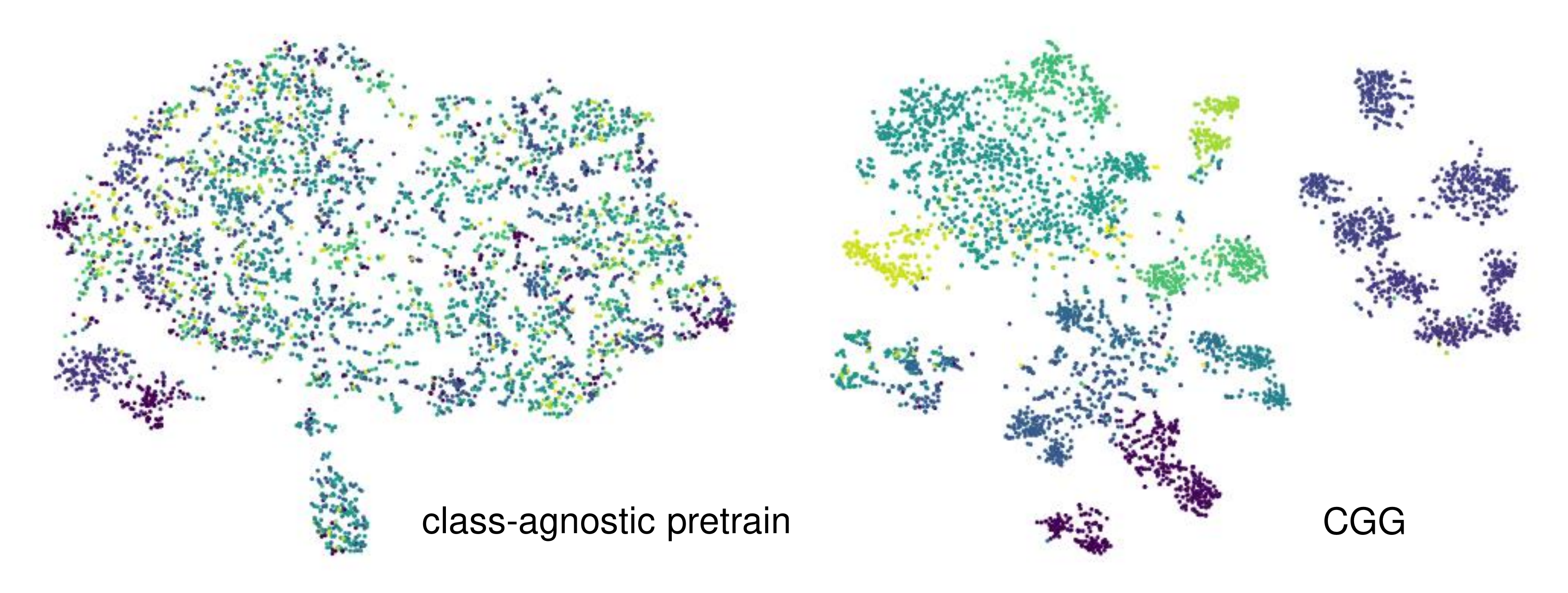}
	\caption{The multi-modal embeddings ${e^M_i}$ in a 2D space using t-SNE \cite{T-SNE}. The colors indicate the class labels of the 17 novel COCO classes. The dots represent the embeddings with the masks that match the ground truth annotations.}
	\label{fig:t-sne.}
\end{figure}

\noindent
\textbf{Layers of Caption Generator.}
We examine the effect of numbers in the Transformer decoder layers in the caption generator. Tab. \ref{tab:ablation_d} shows the results for 2, 4, and 6 layers. 
Adding more layers cannot always boost the performance of novel class AP. 
However, the AP score for all classes increases when the caption generator becomes larger. 
It is because base categories occur more frequently than novel categories, and benefit more from model enlargement.

\noindent
\textbf{Ability of Caption Generation.}
We also explore the generated caption quality, despite this is not our goal for OVIS. 
In Tab.~\ref{tab:ablation-generation-layer}, we observe that adding more Transformer layers in the caption generator cannot improve the model's ability of caption generation. 
However, in Tab.~\ref{tab:ablation-generation-compare}, we train the model with only caption generation supervision and get lower generation scores than the joint training. 
These results indicate that multitask training may also improve the effectiveness of the generation task.

\noindent
\textbf{Ablation on Class-Agnostic Pretraining.}
We investigate the effect of class-agnostic pre-training on our model. 
The class-agnostic model is trained to segment base and potential novel objects before training the multi-modal embeddings and caption generator. 
Tab.~\ref{tab:ablation_e} reports the results of different pre-training strategies. Without class-agnostic pre-training, the mAP on novel classes drops by 5.7\%. 
If fixing Mask2Former in pre-training and only training multi-modal embeddings and caption generator, the mAP on novel classes drops by 2.0\%.
This indicates end-to-end training plays an important role for query-based segmenter.

\noindent
\textbf{GFLOPs and Parameter Analysis.}
CGG introduces a lightweight Transformer decoder as the caption generator during training. 
As shown in \cref{tab:ablation_f}, this increases the number of parameters by 127.7\% in training, while the total GFLOPs increase only by 0.8\%. 
Since text data is much smaller than images under the same batch size, the additional computational cost brought by the caption generator can be ignored. 
The GFLOPs and Parameters during inference are the same as the Mask2Former baseline.

\noindent
\textbf{Segmentation Results Visualization.}
We present some qualitative results of CGG in Fig.~\ref{fig:visual_results.}. The first row shows panoptic results, and the second shows instance results. 
Novel classes are marked with ``*'' and highlighted in the caption. The result demonstrates that our framework can segment and identify base and novel classes. 
We also show the generated comprehensive captions above.

\noindent
\textbf{Embeddings Space Visualization.}
We visualize the multi-modal embeddings learned by CGG and a class-agnostic pretraining baseline using t-SNE in Fig.~\ref{fig:t-sne.}. 
We extract the predicted embeddings for each image in the COCO validation set and match them with ground truth labels by mask similarity. 
The baseline model fails to cluster the embeddings by their categories due to the lack of class-specific knowledge. 
In contrast, CGG leverages caption grounding and generation to learn discriminative embeddings that align with their semantic classes.

\section{Conclusion} This paper presents a joint Caption Grounding and Generation (CGG) framework for instance-level open vocabulary segmentation. The main contributions are: (1) using fine-grained object nouns in captions to improve grounding with object queries. (2) using captions as supervision signals to extract rich information from other words, which helps to identify novel categories. To our knowledge, this paper is the first to unify segmentation and caption generation for open vocabulary learning. The proposed framework achieves significant improvement on OVIS and OSPS and comparable results on OVOD \textit{without} pre-training on large-scale datasets. 

\noindent
\textbf{Limitation and Future Work.} Due to the limited computation resources, we do not pre-train our framework on extra caption datasets. Moreover, we do not use VLMs such as CLIP for distillation or supervision, and we do not experiment on larger scale datasets, like LVIS and Open-Image~\cite{gupta2019lvis,OpenImages}. We will put these as future work. 

\noindent
\textbf{Acknowledgement.} This work is supported by the National Key Research and Development Program of China (No.2020YFB2103402).
We also gratefully acknowledge the support of SenseTime Research for providing the computing resources for this work.


\appendix

\section{More Implementation Details}
\label{sec:implementation}

\noindent
\textbf{Baseline Details.} All the table results in main paper use \textbf{the same ResNet50~\cite{resnet} backbone} for a fair comparison. 
The number of object queries is \textit{100} by default for all experiments.
Our method is trained by only 12 epochs on the COCO training set and evaluated on the COCO validation set. 
All the experiments are carried out on 8 V100 GPUs. Following previous methods~\cite{xpm,zareian2021open}, we use mAP (mean AP on the IoU threshold of 0.5) as the metric for OVIS.

\noindent
\textbf{Training and Inference Details.} We adopt the default training of Mask2Former~\cite{detectron2,cheng2021mask2former,chen2019mmdetection}. A learning rate multiplier of 0.1 is applied to the backbone. For data augmentation, we use the default large-scale jittering (LSJ) augmentation with a random scale sampled from the range 0.1 to 2.0 with the crop size of 1024 $\times$ 1024. We use the default Mask R-CNN inference setting~\cite{maskrcnn}, where we resize an image with a shorter side to 800 and a longer side to 1333. 
\textit{For the inference of OSPS},  we do not use the default joint merge for things and stuff that is used in Mask2Former~\cite{cheng2021mask2former}.
Instead, we put the thing mask first and fill the remaining area with stuff mask prediction. 
In the experiment part, we find that the thing predictions for the unknown are usually in a low score, and they may be covered by high score stuff mask prediction. 
This is because all the stuff masks are trained in a supervised manner.

\noindent
\textbf{Training Splits For OVIS and OSPS.}
For OVIS, we follow the 48/17 split in COCO proposed by \cite{zeroshotobjectdetection}, in which 48 classes are base classes, and 17 are novel classes. For OSPS, we follow the unknown things split proposed by \cite{eopsn}. The unknown percentages are 5\%, 10\%, and 20\% separately.

Concretely, for 48/17 split of OVIS, the \textbf{base} classes are: “person”,
“bicycle”, “car”, “motorcycle”, “truck”, “boat”, “bench”,
“bird”, “horse”, “sheep”, “zebra”, “giraffe”, “backpack”,
“handbag”, “skis”, “kite”, “surfboard”, “bottle”, “spoon”,
“bowl”, “banana”, “apple”, “orange”, “broccoli”, “carrot”, “pizza”, “donut”, “chair”, “bed”, “tv”, “laptop”, “remote”, “microwave”, “oven”, “refrigerator”, “book”,
“clock”, “vase”, “toothbrush”, “train”, “bear”, “suitcase”,
“frisbee”, “fork”, “sandwich”, “toilet”, “mouse”, “toaster”.

The \textbf{novel} classes are: ’bus’, ’dog’, ’cow’, ’elephant’, ’umbrella’, ’tie’, ’skateboard’, ’cup’, ’knife’, ’cake’,
’couch’, ’keyboard’, ’sink’, ’scissors’, ’airplane’, ’cat’, ’snowboard’.

For OSPS, the \textbf{unknown} things are:
5\%: “car”, “cow”, “pizza”, “toilet”.
10\%: “boat”, “tie”, “zebra”, “stop sign”.
20\%: “dining table”, “banana”, “bicycle”, “cake”, “sink”, “cat”, “keyboard”, “bear”.

\noindent
\section{More Experiments Results}
\label{sec:more_exp_results}

\begin{table}[!t]
   \centering
    \caption{ OVR-CNN experiments for Open Vocabulary Object Detection on COCO with 48/17 spilt. ``Vanilla'' means the origin OVR-CNN model without our proposed modules.}
   \scalebox{0.9}{
   \setlength{\tabcolsep}{2.5mm}{\begin{tabular}{c|c|c|c|c|c}
      \toprule[0.15em]
      \multirow{2}{*}{Method} & \multicolumn{2}{c|}{Constrained}  & \multicolumn{3}{c}{Generalized} \\
        \cline{2-6}
        & Base & Novel & Base & Novel & All \\
        \hline
        Vanilla & 40.6 & 22.6 & 39.8 & 18.5 & 34.2 \\
        w. Gro  & 40.3 & 23.3 & 39.4 & 19.6 & 34.2 \\
        w. Gro \& Gen & 40.6 & 23.4 & 40.3 & 18.9 & 34.7 \\
      \bottomrule[0.10em]
   \end{tabular}}}
   \label{tab:result_OVR-CNN}
\end{table}

\noindent
\textbf{Will Joint Grounding and Captioning Help Other Architectures?}
We conduct experiments on a previous model, OVR-CNN~\cite{zareian2021open}, to further evaluate the effectiveness of our proposed modules, i.e., caption grounding with object nouns and caption generation. We re-implement OVR-CNN using PyTorch and add the two modules onto its architecture. The training schedule and results may be different from the original paper~\cite{zareian2021open}, while the training settings are the same in our experiments. Concretely, we train 40,000 steps with a batch size 56 for the caption pre-training stage and 30,000 steps with a batch size 48 for the detector fine-tuning stage. Tab.~\ref{tab:result_OVR-CNN} shows that by adding caption grounding with object nouns, the novel AP score increases, which indicates the effectiveness of our proposed method.
However, adding a caption generation module does not bring further improvement. This may be explained that OCR-CNN already applies ITM and MLM losses as auxiliary losses during the pre-training process, which extracts knowledge from all words in the captions.

\noindent
\textbf{Will Joint Grounding and Captioning Help the Fully Supervised Baseline?} To answer this question, we perform ablation on fully supervised settings in Tab.~\ref{tab:sup_fully}. For the proposed CGG, we verify two main components, including caption grounding and caption generation. Class-emb means only using pre-trained text embeddings for mask classification. Class-label is a traditional learnable, fully connected layer that converts the classes into contiguous labels. In Tab.~\ref{tab:sup_fully}, we observe that the fully supervised method achieves better results than using class embeddings in all three tasks. As shown in the last three rows of Tab.~\ref{tab:sup_fully}, for within-class embedding settings, the added caption grounding and generation modules help to improve the performance on OVIS, but bring no performance gain on OSPS. We conclude that joint grounding and captioning have limited benefits (0.5\% improvements) in supervised settings.

\begin{table}[!t]
   \centering
    \caption{ Ablation on fully supervised instance segmentation, object detection, and panoptic segmentation. AP-novel indicates the mean AP on the 17 novel classes (trained in the fully supervised setting). AP-bbox indicates object detection.}
   \scalebox{0.9}{
   \setlength{\tabcolsep}{1.5mm}{\begin{tabular}{c|c|c|c|c|c|c}
      \toprule[0.15em]
      \multirow{2}{*}{Method} & \multicolumn{3}{c|}{Instance} & \multicolumn{3}{c}{Panoptic} \\
        \cline{2-7}
        & AP & AP-novel & AP-bbox & PQ & PQ-th & PQ-st \\
        \hline
        class-label & 59.3 & 66.6 & 58.9 & 46.4 & 51.9 & 38.2 \\
        class-emb. & 50.6 & 57.8 & 50.2 & 44.4 & 50.5 & 35.1 \\
        w/ gro. & 50.8 & 57.4 & 50.3 & 44.1 & 50.3 & 35.0 \\
        w/ gen. & 50.9 & 57.6 & 50.7 & 44.2 & 50.5 & 34.8 \\
        w/ both. & 51.3 & 57.5 & 50.7 & 44.3 & 50.6 & 34.9 \\
      \bottomrule[0.10em]
   \end{tabular}}}
   \label{tab:sup_fully}
\end{table}

\begin{table}[!t]
   \centering
    \caption{ Ablation on layers of Caption Generator and quality of Open Vocabulary Instance Segmentation. We adopt BLUE, CIDEr, and ROUGE as the metrics to evaluate the quality of generated captions.}
   \scalebox{0.9}{
   \footnotesize
   \setlength{\tabcolsep}{0.45mm}{\begin{tabular}{c|c|c|c|c|c|c|c|c|c}
      \toprule[0.15em]
      \multirow{2}{*}{Layers} & \multicolumn{3}{c|}{Segmentation} & \multicolumn{6}{c}{Caption Generation} \\
        \cline{2-10}
        & Base & Novel & All & BLUE-1 & BLUE-2 & BLUE-3 & BLUE-4 & CIDEr & ROUGE \\
        \hline
        2 & 46.7 & 23.4 & 40.6 & 0.473 & 0.311 & 0.206 & 0.141 & 0.307 & 0.360 \\
        4 & 46.0 & 28.4 & 41.4 & 0.418 & 0.258 & 0.166 & 0.111 & 0.239 & 0.320 \\
        6 & 48.2 & 26.9 & 42.6 & 0.387 & 0.226 & 0.138 & 0.088 & 0.171 & 0.289 \\
      \bottomrule[0.10em]
   \end{tabular}}}
   \label{tab:sup_gen_OVIS}
\end{table}

\noindent
\textbf{Will Better Caption Generator Help Open Vocabulary Instance Segmentation?} We further explore the influence of the caption generation module to open vocabulary instance segmentation. Tab.~\ref{tab:sup_gen_OVIS} shows the results. As we adopt a larger caption generator, the overall segmentation quality (AP all) increases. On the contrary, the quality of the caption (including BLUE and CIDEr) generation drops. This means a better caption generator may not be a better open vocabulary instance segmenter. The role of the caption generator is to force the model to know the existence of novel objects, so pursuing a better caption generation model is not our goal for OVIS and OSPS.

\section{Visual Analysis and Comparison}
\label{sec:visual}

\begin{figure}[!t]
	\centering
	\includegraphics[width=1\linewidth]{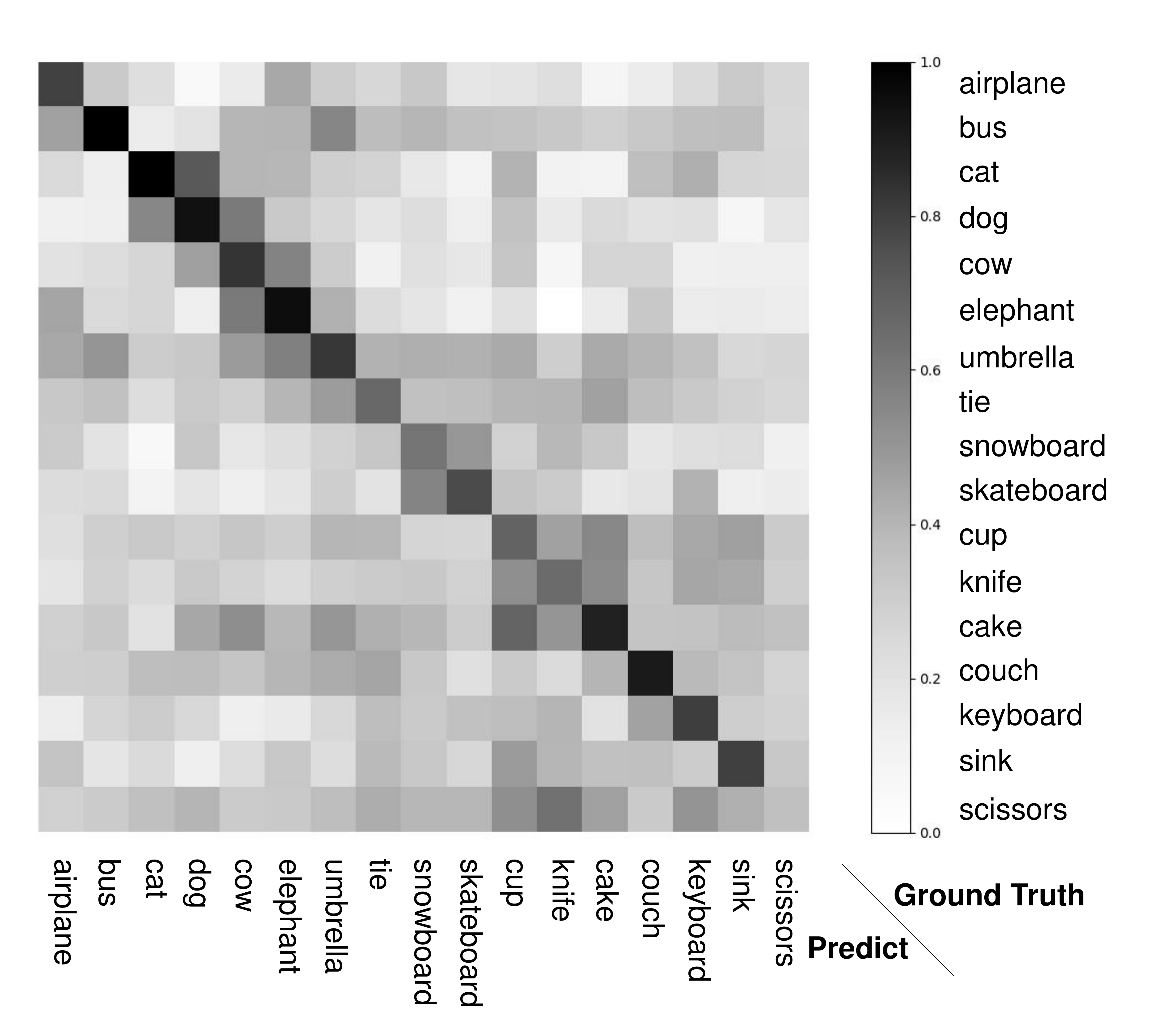}
        \vspace{-8mm}
	\caption{The correlation map between Ground Truth and model predictions on \textbf{novel classes}.
	The noun embeddings and object queries for novel classes are highly correlated.}
	\label{fig:query_cat_cmap}
\end{figure}

\noindent
\textbf{Visualization Analysis both Nouns and Object Queries.} We calculate the correlation map between the predicted multi-modal embeddings $e_i$ and the Ground Truth class embeddings. As shown in Fig. \ref{fig:query_cat_cmap}, our model can correctly distinguish novel classes based on the segmentation masks.

\begin{figure}[!t]
	\centering
	\includegraphics[width=1\linewidth]{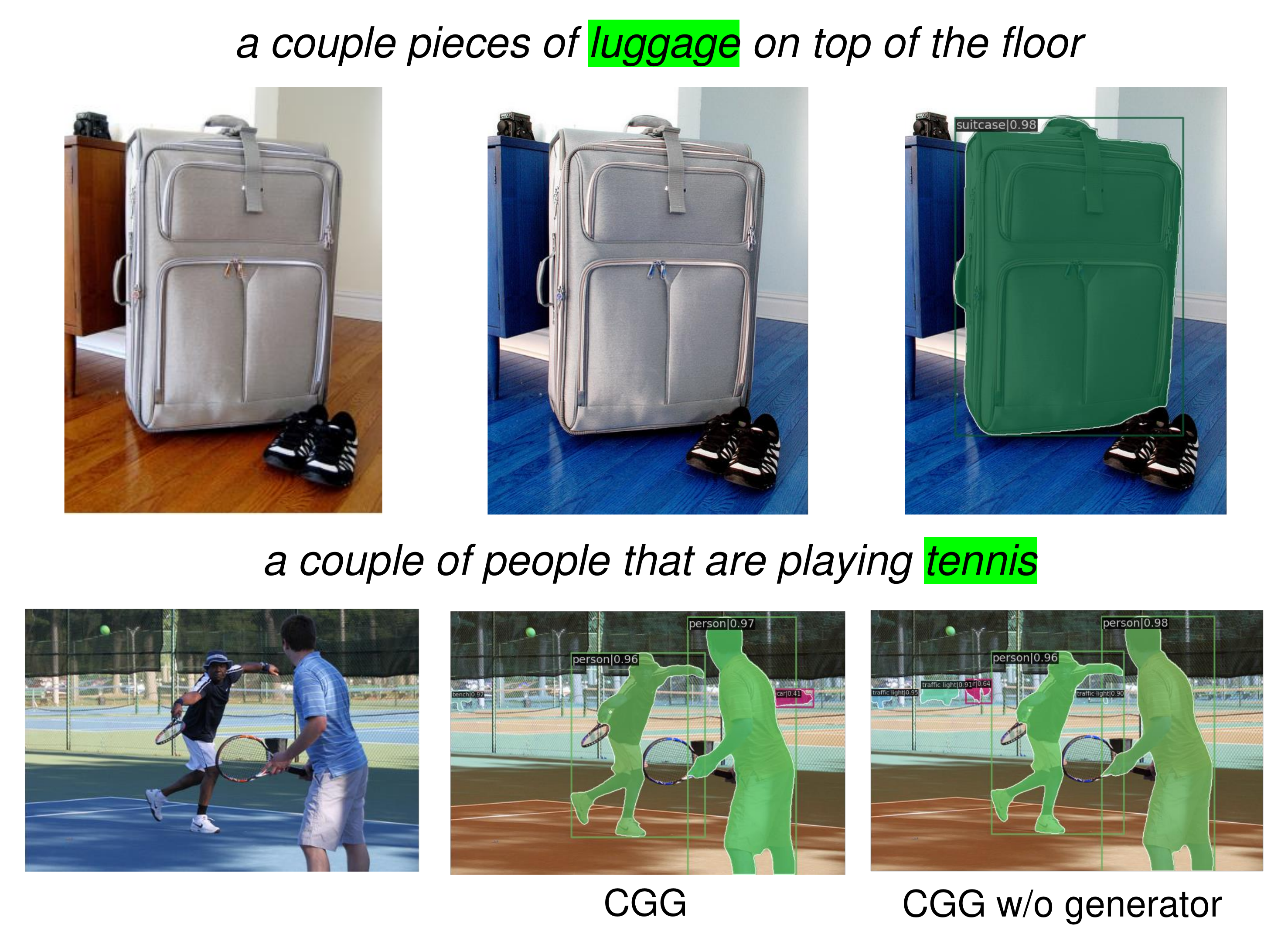}
        \vspace{-8mm}
	\caption{Examples of captions predicting objects that are not in the category list.}
	\label{fig:sup_visual_cap2}
\end{figure}

\begin{figure*}[!t]
	\centering
	\includegraphics[width=1\linewidth]{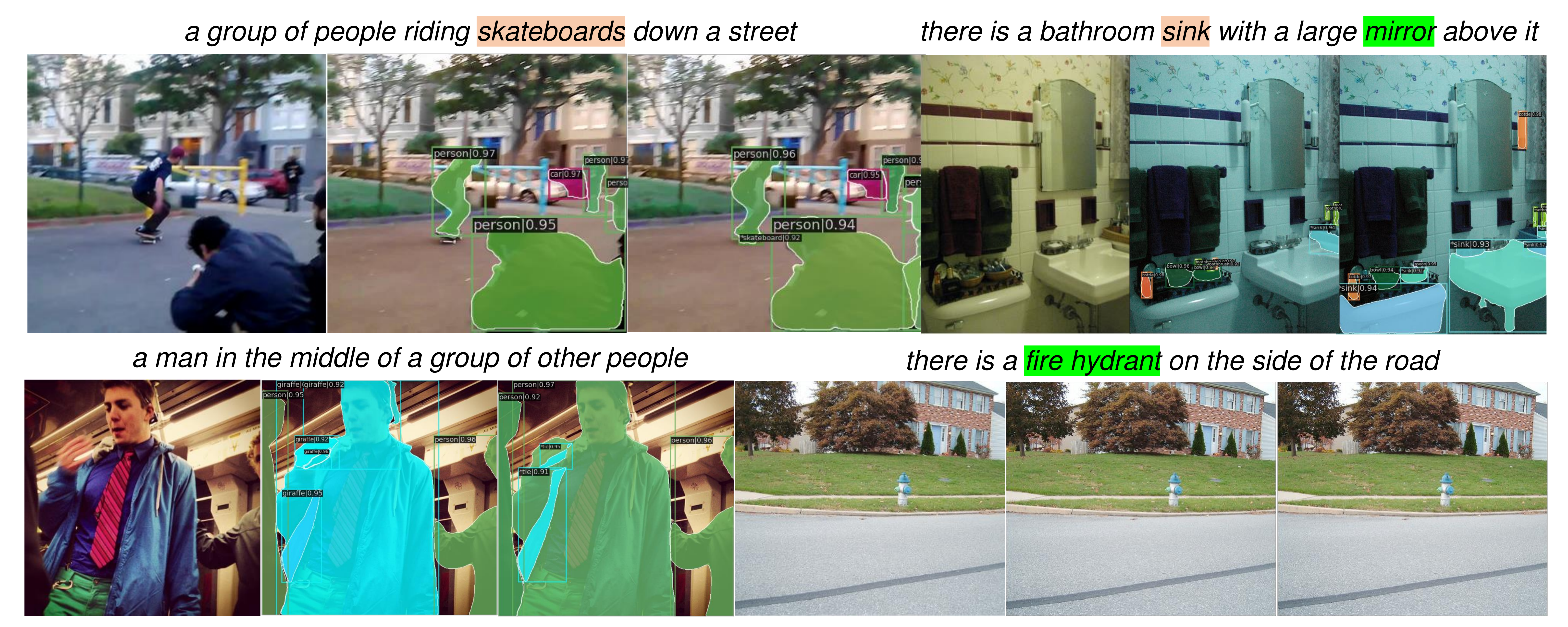}
	\caption{Visualization results of generated captions and the related segmentations of CGG. Input Image (Left), CGG w/o caption generation (Middle), CGG (Right). ``mirror'' and ``fire hydrant'' are not in the category list (both base and novel) but are still mentioned in the generated captions.}
	\label{fig:sup_visual_cap}
\end{figure*}

\begin{figure*}[!t]
	\centering
	\includegraphics[width=1\linewidth]{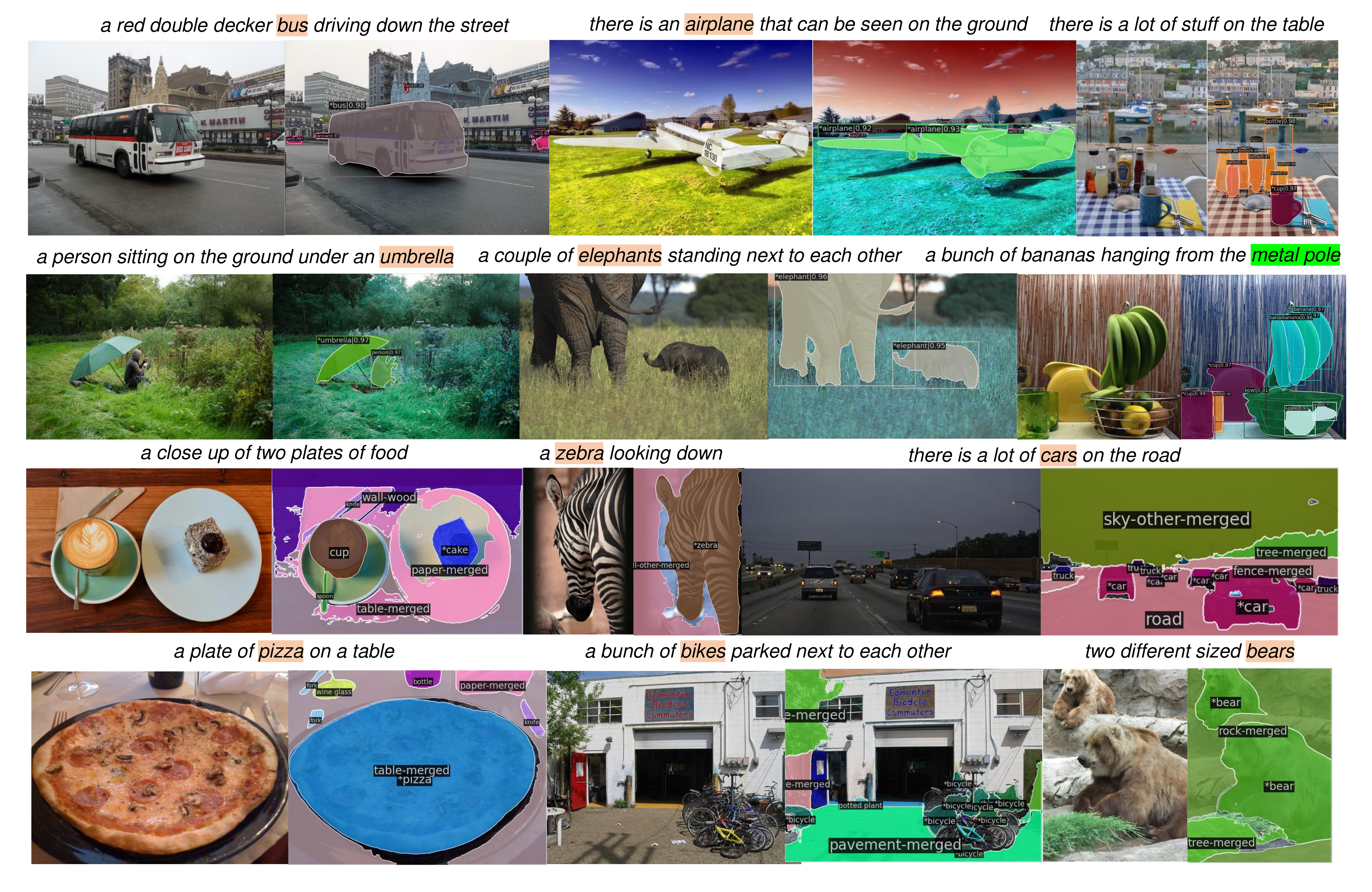}
	\caption{More visualization results of OVIS (Top two rows) and OSPS (Bottom two rows). Novel classes are marked by ``*''.}
	\label{fig:sup_visual}
\end{figure*}

\begin{figure*}[!t]
	\centering
	\includegraphics[width=1\linewidth]{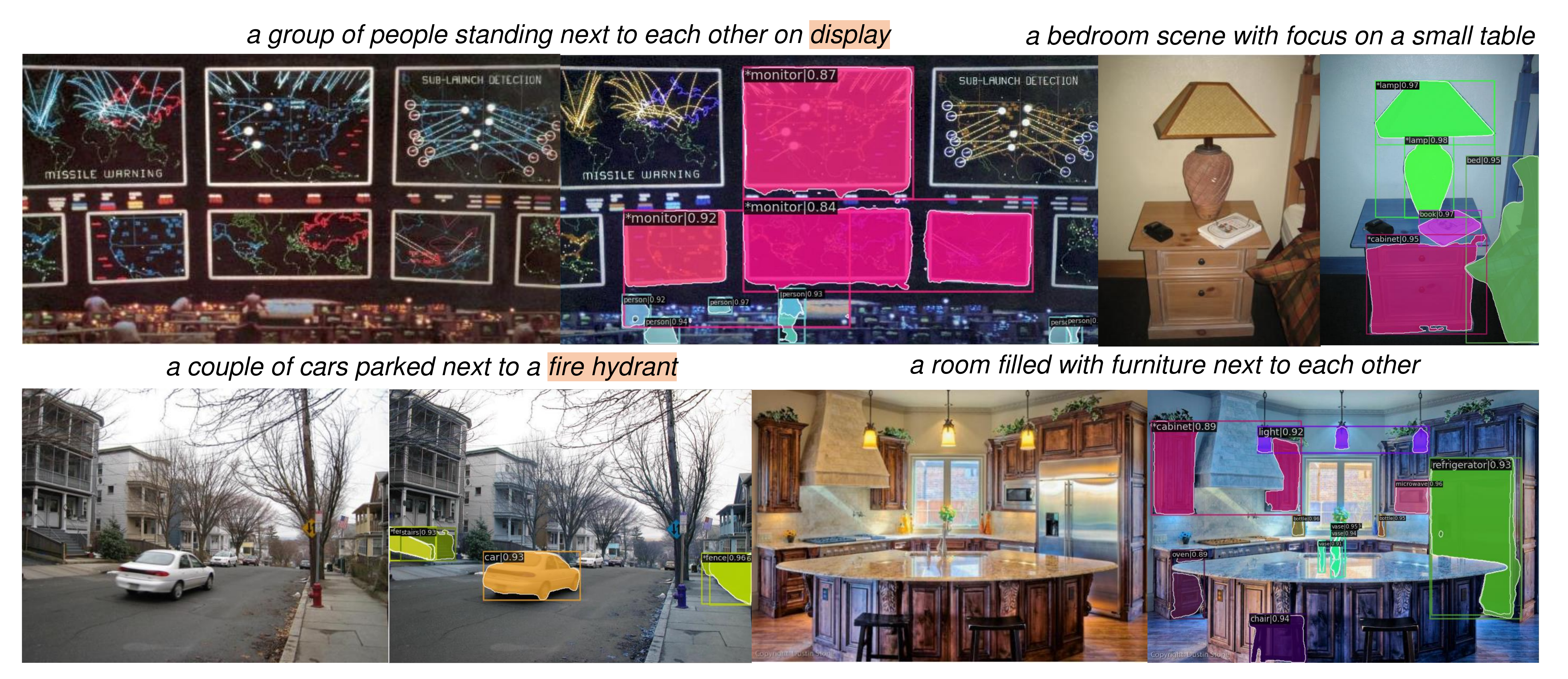}
        \vspace{-8mm}
	\caption{Visualization on ADE20k \cite{ade20k}. Following \cite{cheng2021mask2former}, we apply instance segmentation on 100 instance classes. Classes not in COCO are marked by ``*''.}
	\label{fig:vis_ade20k.}
\end{figure*}

\noindent
\textbf{More Visual Examples from Caption Generation.} We observe that in some cases, the caption generated by CGG can predict objects that are \textit{not} in the category list. 
Categories beyond the given list cannot be correctly classified using the similarity between multi-modal embeddings and class embeddings since the class embeddings are not accessible during inference, like in top images of Fig.~\ref{fig:sup_visual_cap2}. 
There is a couple of luggage on the floor, but ``luggage'' is not a class in the validation dataset. 
\textit{Without} a caption generator, the model classifies the luggage as ``suitcase''. 
However, with the caption generation module, the generated caption successfully depicts the word ``luggage''. 
In the bottom images, ``tennis'' is also described by captions. Fig. \ref{fig:sup_visual_cap} shows more visualization results with captions.

\noindent
\textbf{More Visualization Results on OVIS and OSPS.} In Fig.~\ref{fig:sup_visual}, we present more visual results of OVIS and OSPS tasks. The CGG model can well segment and classify novel categories well.

\noindent
\textbf{Zero Shot Visualization on ADE20K dataset.} In Fig.~\ref{fig:vis_ade20k.}, we show the visualization results on ADE20K dataset~\cite{ade20k}. CGG can detect and segment novel classes in a zero-shot manner on ADE20K. At the same time, CGG generates comprehensive captions that well depict the content of the images.

{\small
\bibliographystyle{ieee_fullname}
\bibliography{egbib}
}

\end{document}